\documentclass[runningheads]{llncs}
\usepackage{multirow} 
\usepackage[T1]{fontenc}
\usepackage{tabularx}
\usepackage{amsmath}
\usepackage{bbold}
\usepackage{dsfont}
\usepackage{boldline}
\usepackage{threeparttable}
\usepackage[ruled,vlined]{algorithm2e}
\usepackage{orcidlink}
\usepackage{graphicx}

\usepackage{parselines} 
\begin{document}
\title{
MAML-KT: Addressing Cold Start Problem in Knowledge Tracing for New Students via Few-Shot Model-Agnostic Meta Learning
}
\titlerunning{Model-Agnostic Meta Learning for Cold Start Knowledge Tracing}
%
\author{Indronil Bhattacharjee \orcidlink{0000-0002-3463-3389} \and
Christabel Wayllace \orcidlink{0000-0001-8039-2777}}
 \authorrunning{I. Bhattacharjee and C. Wayllace}
 \institute{New Mexico State University, Las Cruces, New Mexico, USA \\
 \email{\{indronil, cwayllac\}@nmsu.edu}}
\maketitle
%
\begin{abstract}

Knowledge tracing (KT) models are commonly evaluated by training on early interactions from all students and testing on later responses. While effective for measuring average predictive performance, this evaluation design obscures a cold start scenario that arises in deployment, where models must infer the knowledge state of previously unseen students from only a few initial interactions. Prior studies have shown that under this setting, standard empirically risk-minimized KT models such as DKT, DKVMN and SAKT exhibit substantially lower early accuracy than previously reported. We frame new-student performance prediction as a few-shot learning problem and introduce MAML-KT, a model-agnostic meta learning approach that learns an initialization optimized for rapid adaptation to new students using one or two gradient updates. We evaluate MAML-KT on ASSISTment data using a controlled cold start protocol that trains on a subset of students and tests on held-out learners across early interaction windows, scaling cohort sizes from 10 to 50 students. Across datasets, MAML-KT achieves higher early accuracy than prior KT models in nearly all cold start conditions. Overall, optimizing KT models for rapid adaptation reduces early prediction error
and sharpens the interpretation of early accuracy fluctuations. 

\end{abstract}
\section{Introduction}
Personalized tutoring systems rely on accurate early estimates of a learner’s mastery to decide what to present next and how to adapt difficulty. Knowledge Tracing (KT) models this as sequential prediction over student responses to tutoring items. Modern deep KT approaches, including recurrent models \cite{dkt} and memory-based architectures \cite{skvmn,dkvmn}, are typically trained via empirical risk minimization (ERM). 

While these models capture temporal structure and concept dynamics, they can struggle in cold start settings \cite{indronil,coldstart_skill}. When a new student has only a few interactions, parameters optimized for average performance may not personalize quickly, and early errors can influence subsequent instructional decisions. 
Prior work has formally characterized the new-student cold start problem under disjoint train–test splits, documenting unstable early-phase performance but leaving open the question of how to explicitly mitigate it~\cite{indronil}. To address this limitation, we frame new-student KT as a few-shot adaptation problem and apply Model-Agnostic Meta Learning (MAML) \cite{finn2017}. Rather than optimizing a single global solution, MAML learns an initialization that can be rapidly adapted to a new student from a small support prefix.

Our contributions are threefold: (1) we formulate new-student knowledge tracing as a few-shot adaptation problem under a strictly causal support–query split; (2) we introduce MAML-KT, a model-agnostic meta-learning approach tailored to sequential student data for rapid personalization; and (3) we provide a systematic evaluation of cold-start performance across multiple datasets and cohort sizes (10–50 students), showing that meta-learned initialization improves early-phase prediction and scales to larger, more realistic deployment settings.

\section{Background and Related Works}

Knowledge Tracing \cite{kt} estimates a learner’s latent mastery from interaction sequences to predict future performance. DKT uses RNNs to model interaction histories \cite{dkt}, while DKVMN externalizes concept representations via memory and attention-based models weight relevant past interactions \cite{dkvmn,sakt,akt}. Despite strong average performance, these globally trained models require multiple observations before predictions stabilize for a new student, exposing the new-student cold start problem \cite{indronil}.

Cold start arises when a model must personalize to a new student with only a few interactions or when new skills are introduced. Unlike standard evaluation, the focus is performance over early interactions. Architectural approaches improve early predictions through inductive bias \cite{cskt}, and some methods incorporate auxiliary information to reduce uncertainty \cite{jung,guo:24}, but most KT models remain optimized for global prediction rather than rapid per-student adaptation.

Meta learning trains models to adapt quickly across tasks. In MAML \cite{finn2017}, parameters are optimized so that a few gradient steps on a support set yield strong query performance. This paradigm has been effective in cold-start recommendation settings, where each user defines a task with sparse interactions~\cite{lu2020,du2022,wang2023}. 

\section{Problem Statement and Research Approach}
\subsection{Problem Statement: New-Student Cold Start in KT}

We consider knowledge tracing (KT) in the new-student cold start setting \cite{indronil}, where an unseen student has no prior history and the model must predict correctness on upcoming items using only the first $K$ interactions. Our goal is to learn parameters that can be quickly personalized via a few gradient steps on these initial interactions. 

Let a student’s sequence be $S_s={(q_t,a_t)}_{t=1}^{T_s}$, where $q_t$ denotes the question (with one or more associated skills) and $a_t \in {0,1}$ its correctness label. For each unseen student $s$, we define a causal split:
$S_s^{support}={(q_t,a_t)}_{t=1}^K$ and
$S_s^{query}={(q_t,a_t)}_{t=K+1}^{T_s}$. 
 
We evaluate next-step correctness on the query segment, emphasizing early-phase performance for small $K$. 

\subsection{Key Research Questions}
Our study aims to address the following research questions:
\begin{enumerate}
    \item Does meta learning improve early-phase new student performance over ERM baselines (DKT, DKVMN, SAKT) at small $K$?
    \item Does the proposed MAML-KT approach scale with cohort size, i.e., do its cold start accuracy change when moving from the prior settings (10 students) to larger cohorts (20 and 50 students)?
    \item
    Under which sequence and content conditions does MAML-KT trail other ERM baselines?
    
\end{enumerate}

\subsection{Few-Shot Task Construction}
For each student trajectory $\{(q_t, a_t)\}_{t=1}^T$, we construct a few-shot task under next-step prediction. 
At timestep $t$, the model receives the history token $x_t=(q_t,a_t)$ and predicts the subsequent outcome $a_{t+1}$ conditioned on the target item $q_{t+1}$. 

Thus, per-timestep training examples are $(x_t, q_{t+1}, a_{t+1})$ for $t=1,~\dots,T-1$.
Given a support size $K$, we split each sequence into a causal support prefix and query suffix. 
The support set consists of the first $K$ timesteps, and the query set consists of the remaining timesteps. 
To ensure a non-empty query, we require $1 \le K \le T-1$ and discard sequences with $T<2$. 
Padding is applied during preprocessing and does not affect interaction order.

\subsection{Meta learning Objective}

Following the MAML paradigm~\cite{finn2017}, the meta-parameter $\theta_{\mathcal{T}}$ is optimized such that after inner-loop adaptation on the support set of a student task $\mathcal{T}$, the adapted parameters $\theta'_{\mathcal{T}}$ minimize the query loss with inner learning rate $\alpha$: 
\begin{equation*}
\min_{\theta} \mathbb{E}_{\mathcal{T} \sim p(\mathcal{T})} 
[ \mathcal{L}_{\mathcal{T}}^{\text{query}}\!(\theta'_{\mathcal{T}}) ]\quad  \text{s.t. \quad} \theta'_{\mathcal{T}} = \theta - \alpha \nabla_{\theta} \mathcal{L}_{\mathcal{T}}^{\text{support}}(\theta) \end{equation*}
Here, $\theta$ denotes the shared initialization parameters, $\theta'_{\mathcal{T}}$ the task-adapted parameters for student $\mathcal{T}$, and $\mathcal{L}$ is the binary cross-entropy loss over next-step prediction.

\subsection{Training Procedure}

During meta-training, we iterate over meta-batches of student tasks. 
For each student $s$ in the batch:

\noindent\textbf{Support adaptation.} For each student, we update the shared parameters using gradient descent on the support loss $\mathcal{L}_{support}^{(s)}$. The update is differentiable, enabling the outer meta-optimization to account for how the model adapts to new students.

\noindent\textbf{Meta-loss on query.} Using the adapted parameters $\theta'_s$, we compute the query loss $\mathcal{L}_{query}^{(s)}$ on the remaining timesteps. These per-task query losses are averaged across the meta-batch to form the meta-objective.

\noindent\textbf{Outer update.} We backpropagate through the inner updates and update the shared initialization,
$\theta \leftarrow \theta - \beta \nabla_{\theta} \left( \frac{1}{B} \sum_{s=1}^{B} \mathcal{L}_{\text{query}}^{(s)} \right)$
using Adam with meta learning rate $\beta$. 

\subsection{Evaluation Protocol}

Let $N$ denote the total number of query predictions aggregated across all test students. We calculate
Overall accuracy and Windowed early-phase accuracy.

\begin{equation}
\mathrm{ACC}
= \frac{1}{N}\sum_{t=1}^{N} \!\left( \mathbf{1}\!\left[\hat{p}_t \ge 0.5\right] = y_t \right)
\label{eqn:acc}
\end{equation}

For a window $[Q_{min}, Q_{max}]$ and $\text{coldstart zone} \in \{\text{Critical}, \text{Moderate}\}$, the average windowed accuracy is

\begin{equation}
\overline{ACC}_{\text{coldstart zone}}=\frac{1}{N_Q}\sum_{Q=Q_{min}}^{Q_{max}}ACC(Q)
\label{eqn:avg acc}
\end{equation}

\subsection{Algorithm}
We retain the standard MAML objective \cite{finn2017}, optimizing query loss after inner adaptation, and adapt it to sequential student–response data (Algorithm~\ref{alg:maml-kt}). Each student trajectory defines a task. 

The backbone is a GRU-based Deep KT model with a projected target-item embedding fused before the readout layer. Inner adaptation performs $R$ steps of task-local SGD on the support loss. We train second-order MAML by backpropagating through the inner updates and applying the meta-update to the averaged query loss across tasks.

The algorithm has the same meta-objective and inner/outer optimization as standard MAML. The differences are limited to task construction and model design: a causal support–query split with auto-shrink, sequence-aware preprocessing, and a KT-specific GRU with target fusion. 
The objective and meta-optimization remain unchanged.
The code is available at \textbf{github.com/Indronil-Prince/MAML-KT}.

\begin{algorithm}[t]
\LinesNumbered
\SetAlgoLined
\caption{MAML-KT (GRU backbone, second-order)}
\label{alg:maml-kt}
\KwIn{Training set $\mathcal{D}_{train}$, support size $K$, inner steps $R$, LR $\alpha$, meta LR $\beta$ meta-batch $B$
}

\textbf{Preprocess}: For each student trajectory $\{(q_t,a_t)\}_{t=1}^T$, 
form next-step pairs $(x_t,q_{t+1},a_{t+1})$ with $x_t=(q_t,a_t)$; 
discard $T<2$ and enforce $1 \le K \le T-1$.

\textbf{Model}: GRU over interaction tokens with projected target-item embedding.

\For{epoch $=1,2,\dots$}{
  Sample meta-batch $\{(X_i,y_i,T_i)\}_{i=1}^B$ from $\mathcal{D}_{train}$\;
  $L_{\text{meta}} \leftarrow 0$, $V \leftarrow 0$\;
  
  \For{$i=1$ \KwTo $B$}{
    \textbf{if} $T_i \le K$ \textbf{continue}
    
    support $(X^s,y^s) = (X_i[1{:}K], y_i[1{:}K])$\;
    query $(X^q,y^q) = (X_i[K{+}1{:}T_i-1], y_i[K{+}1{:}T_i-1])$ \tcp*{Causal split}
    
    $\phi \leftarrow \theta$ \tcp*{fast parameters}
    
    \For{$r=1$ \KwTo $R$}{
      $\ell_r \leftarrow \mathrm{BCE}\big(f(X^s;\phi),\, y^s\big)$\;
      $\phi \leftarrow \phi - \alpha \nabla_\phi \ell_r$\;
    }
    
    $\ell_q \leftarrow \mathrm{BCE}\big(f(X^q;\phi),\, y^q\big)$\;
    $L_{\text{meta}} \leftarrow L_{\text{meta}} + \ell_q$; $V \leftarrow V+1$\;
  }
  
  \textbf{if} {$V>0$,} $\textbf{then} \: \theta \leftarrow \theta - \beta \nabla_\theta (L_{\text{meta}}/V)$\;
}
\end{algorithm}

\section{Experiment setup and methodology}

\subsection{Datasets}
We use three ASSISTments benchmarks: ASSIST2009 Skill-Builder \cite{assist2009}, ASSIST2015 \cite{assist2015}, and ASSIST2017 Challenge \cite{assist2017}. These datasets contain student-problem interactions from mathematics curricula, including question IDs, binary correctness labels and question–skill mappings. 

\subsection{Data Segregation and Problem Setup}

We follow Bhattacharjee et al. (2025) \cite{indronil}, applying a minimum-length filter before sampling: students must have $\ge 20$ interactions in ASSIST2009 and ASSIST2015, and $\ge 30$ in ASSIST2017. From each filtered dataset, we adopt the same new-student protocol used in \cite{indronil} for cohort size 10. We additionally construct cohorts of 20 and 50 students via uniform random sampling of student IDs. We define the critical (Q=3–10) and moderate (Q=11–15) cold-start windows following prior work, corresponding to phases where limited interaction history constrains personalization and where early instructional decisions are most impactful~ \cite{indronil}. 

For each dataset and cohort size, we generate five independent splits. We frame meta-training as per-student tasks. Each training student's sequence is split chronologically into a support prefix and query suffix, one fast gradient update on the support adapts the KT backbone and the adapted model is evaluated on the query segment to refine the shared initialization. No cross-student leakage is allowed. 

At test time, each held-out student's earliest interactions are used once for adaptation; subsequent interactions are predicted with the adapted weights and no further learning occurs.

Hyperparameters, including learning rates ($\alpha$, $\beta$), number of inner-loop steps, and hidden dimensions, were selected via validation on training students to maximize early-phase accuracy.
\section{Results and Discussion}
We analyze results along three dimensions: (1) early-phase accuracy (lift-off), (2) stability under limited history, and (3) sensitivity to skill transitions. To isolate the effect of meta-learning, we compare MAML-KT against its ERM counterpart (DKT), which shares the same GRU backbone but is trained without task-level adaptation, as well as standard KT baselines (DKVMN and SAKT).

\begin{table}[t]
\centering
\caption{Critical (Q=3-10) and Moderate cold start (Q=11–15): best (first row) and second-best (second row) accuracies per dataset $\times$ set $\times$ cohort size.}
\label{tab:t1}
\begin{center}
\def\arraystretch{1.4}
\resizebox{\textwidth}{!}{
\tiny{}
\begin{tabular}
{|c|c|c|c|c|cV{4}c|c|c|c|c|}
\hline
\multicolumn{6}{|cV{4}}{\textbf{Critical Cold Start}} & \multicolumn{5}{c|}{\textbf{Moderate Cold Start}}\\
\hline
\tiny\textbf{Dataset} & \tiny\textbf{Set 1} & 
\tiny\textbf{Set 2} & 
\tiny\textbf{Set 3} & 
\tiny\textbf{Set 4} & 
\tiny\textbf{Set 5} & 
\tiny\textbf{Set 1} & 
\tiny\textbf{Set 2} & 
\tiny\textbf{Set 3} & 
\tiny\textbf{Set 4} & 
\tiny\textbf{Set 5} \\ \hline

 \multicolumn{11}{|c|}{\tiny\textbf{10 New Students}}\\ \hline
 
\tiny ASSIST &
\tiny\textbf{75.3$^{ M}$} & 
\tiny\textbf{75.9$^{ M}$} & 
\tiny\textbf{72.3$^{ M}$} & 
\tiny\textbf{67.2$^{ M}$} & 
\tiny\textbf{66.8$^{ M}$} &
\tiny\textbf{79.4$^{ M}$} & 
\tiny\textbf{78.1$^{ M}$} & 
\tiny\textbf{76.6$^{ M}$} & 
\tiny\textbf{71.9$^{ M}$} & 
\tiny\textbf{70.5$^{ M}$}  \\
   
\tiny2009 & 
\tiny{72.0$^{ D}$} & 
\tiny{68.1$^{ S}$} &  
\tiny{68.8$^{ S}$} & 
\tiny{63.9$^{ S}$} & 
\tiny{61.6$^{ S}$} & 
\tiny{76.6$^{ D}$} & 
\tiny{73.3$^{ S}$} & 
\tiny{72.6$^{ S}$} &  
\tiny{70.4$^{ S}$} & 
\tiny{64.7$^{ D}$} \\ 

\tiny ASSIST &
  \tiny\textbf{84.5$^{ M}$} & 
\tiny\textbf{78.1$^{ M}$} & 
\tiny\textbf{70.0$^{ M}$} & 
\tiny\textbf{72.1$^{ M}$} & 
\tiny\textbf{66.0$^{ M}$} &
\tiny\textbf{88.3$^{ M}$} & 
\tiny\textbf{78.7$^{ M}$} & 
\tiny\textbf{76.8$^{ M}$} & 
\tiny\textbf{77.9$^{ M}$} & 
\tiny\textbf{75.6$^{ M}$} \\
   
\tiny 2015 & 
\tiny {76.1$^{ N}$} & 
\tiny{69.8$^{ D}$} & 
\tiny{61.2$^{ S}$} & 
\tiny{65.7$^{ S}$} & 
\tiny{59.9$^{ D}$} &
\tiny{81.0$^{ N}$} & 
\tiny{78.5$^{ N}$} & 
\tiny{71.2$^{ S}$} & 
\tiny{74.9$^{ N}$} & 
\tiny{69.7$^{ N}$}  \\ 

\tiny ASSIST &
\tiny\textbf{67.4$^{ M}$} & 
\tiny\textbf{71.9$^{ M}$} & 
\tiny\textbf{72.9$^{ M}$} & 
\tiny\textbf{70.6$^{ M}$} & 
\tiny\textbf{71.7$^{ M}$} &
\tiny\textbf{69.1$^{ M}$} & 
\tiny\textbf{72.8$^{ M}$} & 
\tiny\textbf{69.4$^{ M}$} & 
\tiny\textbf{72.7$^{ M}$} & 
\tiny\textbf{73.0$^{ M}$} \\
  
\tiny  2017 & 
\tiny{62.7$^{ S}$} & 
\tiny{69.9$^{ S}$} & 
\tiny{66.5$^{ S}$} & 
\tiny{68.2$^{ S}$} & 
\tiny{66.3$^{ S}$} &
\tiny{68.9$^{ S}$} & 
\tiny{71.4$^{ S}$} & 
\tiny{69.0$^{ S}$} & 
\tiny{70.2$^{ S}$} & 
\tiny{69.7$^{ S}$} 
  
  \\ \hline

 \multicolumn{11}{|c|}{\tiny\textbf{20 New Students}}\\ \hline
 
\tiny ASSIST &
\tiny\textbf{81.1$^{ M}$} & 
\tiny\textbf{76.8$^{ M}$} & 
\tiny\textbf{81.0$^{ M}$} & 
\tiny\textbf{75.5$^{ M}$} & 
\tiny\textbf{81.9$^{ M}$} &
\tiny\textbf{82.6$^{ M}$} & 
\tiny\textbf{80.9$^{ M}$} & 
\tiny\textbf{82.7$^{ M}$} & 
\tiny\textbf{82.0$^{ M}$} & 
\tiny\textbf{84.6$^{ M}$}  \\
   
\tiny2009 & 
\tiny{70.8$^{ S}$} & 
\tiny{76.2$^{ S}$} & 
\tiny{76.9$^{ S}$} & 
\tiny{74.0$^{ S}$} & 
\tiny{80.3$^{ S}$} &
\tiny{77.9$^{ S}$} & 
\tiny{80.5$^{ S}$} & 
\tiny{78.0$^{ S}$} & 
\tiny{79.1$^{ S}$} & 
\tiny{81.8$^{ S}$}  \\ 

\tiny ASSIST &
\tiny\textbf{73.6$^{ M}$} & 
\tiny\textbf{77.3$^{ M}$} & 
\tiny{79.5$^{ S}$} & 
\tiny\textbf{77.3$^{ M}$} & 
\tiny\textbf{76.0$^{ M}$} &
\tiny\textbf{76.8$^{ M}$} & 
\tiny\textbf{80.7$^{ M}$} & 
\tiny\textbf{78.0$^{ M}$} & 
\tiny\textbf{81.7$^{ M}$} & 
\tiny\textbf{80.1$^{ M}$} \\
   
\tiny 2015 & 
\tiny{73.4$^{ S}$} & 
\tiny{76.8$^{ S}$} & 
\tiny\textbf{{75.9$^{ M}$}} & 
\tiny{75.9$^{ S}$} & 
\tiny{74.7$^{ S}$} &
\tiny{75.3$^{ D}$} & 
\tiny{79.1$^{ D}$} & 
\tiny{76.1$^{ S}$} & 
\tiny{74.3$^{ D}$} & 
\tiny{73.2$^{ S}$}  \\ 

\tiny ASSIST &
\tiny\textbf{75.7$^{ M}$} & 
\tiny{75.8$^{ S}$} & 
\tiny{76.3$^{ S}$} & 
\tiny\textbf{65.1$^{ M}$} & 
\tiny\textbf{67.6$^{ M}$} &
\tiny\textbf{72.5$^{ M}$} & 
\tiny{69.9$^{ S}$} & 
\tiny{71.7$^{ S}$} & 
\tiny\textbf{78.4$^{ M}$} & 
\tiny\textbf{73.8$^{ M}$}  \\
  
\tiny  2017 & 
\tiny{71.3$^{ S}$} & 
\tiny\textbf{{72.1$^{ M}$}} & 
\tiny\textbf{{71.9$^{ M}$}} & 
\tiny{74.7$^{ D}$} & 
\tiny{74.2$^{ S}$} &
\tiny{68.3$^{_S}$} & 
\tiny\textbf{68.8$^{ M}$} & 
\tiny\textbf{70.3$^{ M}$} & 
\tiny{73.9$^{ D}$} & 
\tiny{72.0$^{ D}$} 
  
  \\ \hline
  
   \multicolumn{11}{|c|}{\tiny\textbf{50 New Students}}\\ \hline
 
\tiny ASSIST &
\tiny\textbf{80.1$^{ M}$} & 
\tiny\textbf{77.8$^{ M}$} & 
\tiny\textbf{74.4$^{ M}$} & 
\tiny\textbf{77.1$^{ M}$} & 
\tiny\textbf{77.5$^{ M}$} &
\tiny\textbf{85.0$^{ M}$} & 
\tiny\textbf{78.4$^{ M}$} & 
\tiny\textbf{72.0$^{ M}$} & 
\tiny\textbf{79.6$^{ M}$} & 
\tiny\textbf{77.2$^{ M}$}  \\
   
\tiny 2009 & 
\tiny{77.9$^{ S}$} & 
\tiny{76.7$^{ S}$} & 
\tiny{71.7$^{ S}$} & 
\tiny{73.9$^{ S}$} & 
\tiny{76.7$^{ S}$} &
\tiny{81.3$^{ S}$} & 
\tiny{77.9$^{ S}$} & 
\tiny{70.9$^{ S}$} & 
\tiny{79.3$^{ S}$} & 
\tiny{75.5$^{ S}$}  \\ 

\tiny ASSIST &
\tiny\textbf{78.2$^{ M}$} & 
\tiny\textbf{79.9$^{ M}$} & 
\tiny\textbf{79.3$^{ M}$} & 
\tiny\textbf{77.2$^{ M}$} & 
\tiny\textbf{81.5$^{ M}$} &
\tiny\textbf{79.7$^{ M}$} & 
\tiny{81.8$^{ S}$} & 
\tiny\textbf{82.6$^{ M}$} & 
\tiny\textbf{80.1$^{ M}$} & 
\tiny\textbf{80.2$^{ M}$} \\
   
\tiny 2015 & 
\tiny{76.5$^{ S}$} & 
\tiny{78.9$^{ D}$} & 
\tiny{78.6$^{ S}$} & 
\tiny{76.3$^{ S}$} & 
\tiny{78.3$^{ D}$} &
\tiny{79.4$^{ S}$} & 
\tiny\textbf{81.4$^{ M}$} & 
\tiny{81.0$^{ S}$} & 
\tiny{77.8$^{ S}$} & 
\tiny{76.8$^{ S}$}  \\ 

\tiny ASSIST &
  \tiny\textbf{71.9$^{ M}$} & 
\tiny\textbf{67.6$^{ M}$} & 
\tiny\textbf{74.0$^{ M}$} & 
\tiny\textbf{65.1$^{ M}$} & 
\tiny\textbf{67.6$^{ M}$} &
\tiny\textbf{71.4$^{ M}$} & 
\tiny\textbf{69.7$^{ M}$} & 
\tiny\textbf{73.0$^{ M}$} & 
\tiny\textbf{65.6$^{ M}$} & 
\tiny\textbf{68.1$^{ M}$} \\
  
\tiny  2017 & 
\tiny{71.2$^{ N}$} & 
\tiny{66.6$^{ D}$} & 
\tiny{71.5$^{ N}$} & 
\tiny{64.9$^{ N}$} & 
\tiny{66.1$^{ N}$} &
\tiny{69.6$^{ N}$} & 
\tiny{67.1$^{ D}$} & 
\tiny{69.6$^{ D}$} & 
\tiny{64.9$^{ N}$} & 
\tiny{65.0$^{ D}$} 
  
  \\ \hline

\end{tabular}}
\end{center}
\flushright\scriptsize\textit{{* M: MAML, D: DKT, N:DKVMN, S: SAKT}}
\end{table}

\subsection{Results on 20 and 50 New Student Cohorts}

We evaluate cold-start performance on larger held-out cohorts of 20 and 50 students to assess whether meta-learned adaptation scales beyond prior small-cohort settings. 

Across datasets (Table \ref{tab:t1}), two consistent patterns emerge. 
1) MAML-KT maintains higher early-phase accuracy than ERM baselines in both the critical (Q=3–10) and moderate (Q=11–15) windows. 2) This advantage remains stable as cohort size increases, indicating that the learned initialization generalizes across larger and more diverse student populations rather than overfitting to small evaluation sets.

Compared to prior work limited to cohorts of 10 students~\cite{indronil}, these results show that meta-learned adaptation produces consistent gains under more realistic deployment conditions, where models must generalize to many unseen learners simultaneously.

\begin{figure}[t]
    \centering
    \includegraphics[width=1\linewidth]{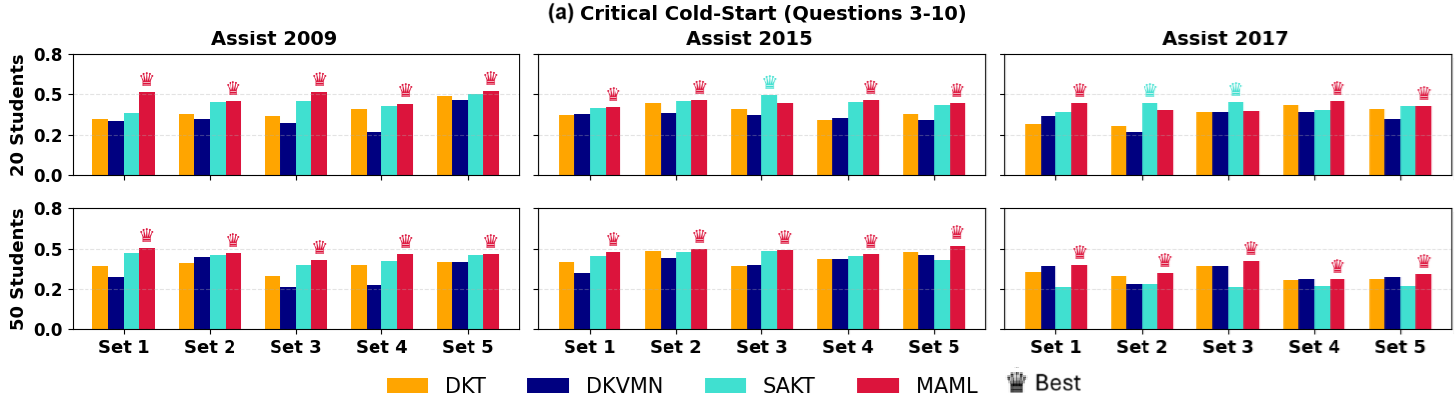}
    \includegraphics[width=1\linewidth]{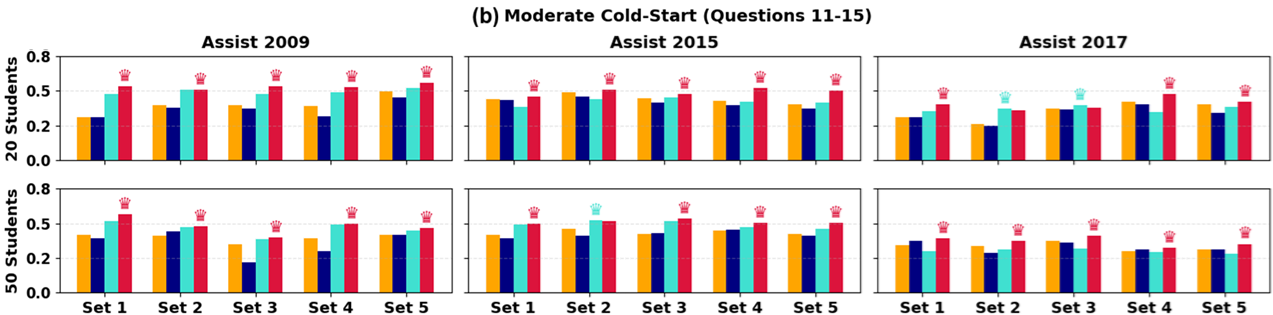}
    \caption{(a) Critical (Questions 3-10) and (b) Moderate Cold Start (Questions 11-15): Average Accuracy across 5 Datasets $\times$ 4 Models $\times$ 2 Cohort Sizes (20 and 50)}
    \label{fig:critical}
\end{figure}

\subsection{Cold Start Performance}

Beyond cohort scaling, we examine how models behave across interaction sequences. Two consistent patterns emerge: (1) faster lift-off, where MAML-KT reaches stable accuracy earlier, and (2) improved stability under limited history, with smoother trajectories across student sets.

On ASSIST2015, these gains are more pronounced despite weaker KC signals, suggesting that task-level adaptation compensates for limited item structure. On ASSIST2017, MAML-KT maintains strong early performance despite greater skill heterogeneity.

These results suggest that training models for rapid adaptation, rather than a single global optimum, better matches the early-stage personalization requirements of tutoring systems.

\subsection{When does MAML-KT trail?}

\begin{figure}[t]
    \centering
    \includegraphics[width=1.0\linewidth]{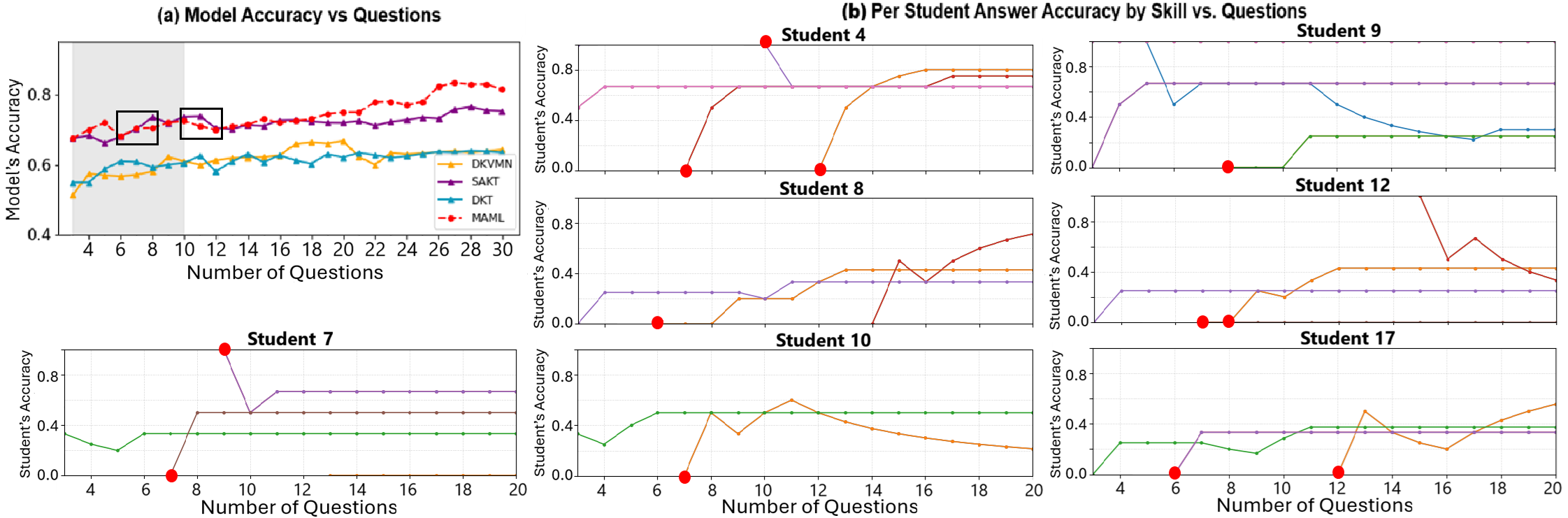}
    \caption{Assist2017 - 20 New Students - Set 2, Questions 6-8 and 10-12 . \\ (a) Model Accuracy vs Questions (b) Per Student Answer Accuracy by Skill vs Questions (The lines represent a skill and start of new skills are marked with red circles)}
    \label{fig:2017-per-student}
\end{figure}

On ASSIST2017, we observe a localized dip around $Q=8$ where MAML-KT briefly trails SAKT before recovering by $Q=13$. Per-student panels (Fig.~\ref{fig:2017-per-student}(b)) show that many learners encounter new skills around $Q=6$–$8$.

Because MAML-KT adapts on the $K$-step support, it specializes to seen skills; when the query introduces unseen skills, performance temporarily drops. In contrast, SAKT does not adapt per student and is less sensitive to this mismatch.

This effect is strongest on ASSIST2017 due to frequent early skill introductions, highlighting a boundary of meta-learning in KT: adaptation relies on short-term skill continuity. When new skills appear, the model effectively faces a form of skill-level cold start. This suggests that early prediction performance is shaped not only by model adaptation capacity, but also by the structure of students’ learning trajectories, particularly the timing and diversity of skill exposure.

\section{Conclusion and Future Work}

We studied MAML for cold-start knowledge tracing by framing each new student as a few-shot adaptation task. Across datasets and cold-start regimes, MAML-KT improved early-phase prediction over ERM baselines, demonstrating that a shared initialization can enable rapid personalization from limited interactions.

Our analysis also reveals an important limitation: gains depend on short-term skill continuity and diminish when new skills appear in the query, highlighting an interaction between student-level and skill-level cold start.

While we instantiate MAML-KT using a GRU-based backbone for comparability with prior KT work, the formulation is model-agnostic and can be extended to other KT architectures.

Future work will investigate adaptation strategies that are more robust to skill shifts, including skill-level task construction and uncertainty-aware updates, toward more reliable and scalable personalization in real instructional settings.

\bibliographystyle{splncs04}
\bibliography{mybibliography}
\end{document}